\documentclass[a4paper, 10pt, conference]{ieeeconf}      % Use this line for a4
\usepackage{FG2017}
\usepackage{amsmath}
\usepackage{graphicx}
\usepackage{cite}
\usepackage{dirtytalk}
\FGfinalcopy % *** Uncomment this line for the final submission

\IEEEoverridecommandlockouts                              % This command is only
\title{\LARGE \bf
FaceNet2ExpNet: Regularizing a Deep Face Recognition Net for Expression Recognition
}

\author{\parbox{16cm}{\centering
    {\large Hui Ding$^1$, Shaohua Kevin Zhou$^2$ and Rama Chellappa$^1$}\\
    {\normalsize
    $^{1}$ University of Maryland, College Park\\
    $^2$ Siemens Healthcare Technology Center, Princeton, New Jersey}}
}

\begin{document}

\ifFGfinal
\thispagestyle{empty}
\pagestyle{empty}
\else
%\author{Anonymous FG 2017 submission\\-- DO NOT DISTRIBUTE --\\}
\pagestyle{plain}
\fi
\maketitle

%%%%%%%%%%%%%%%%%%%%%%%%%%%%%%%%%%%%%%%%%%%%%%%%%%%%%%%%%%%%%%%%%%%%%%%%%%%%%%%%
\begin{abstract}
Relatively small data sets available for expression recognition research make the training of deep networks  for expression recognition very challenging. Although fine-tuning can partially alleviate the issue, the performance is still below acceptable levels as the deep features probably contain redundant information from the pre-trained domain.
In this paper, we present \emph{FaceNet2ExpNet}, a novel idea to train an expression recognition network based on static images. We first propose a new distribution function to model the high-level neurons of the expression network. 
%where the mean is the output of a face net.
Based on this, a two-stage training algorithm is carefully designed. In the pre-training stage, we train the convolutional layers of the expression net, regularized by the face net; In the refining stage, we append fully-connected layers to the pre-trained convolutional layers and train the whole network jointly.
Visualization shows that the model trained with our method captures improved high-level expression semantics.
Evaluations on four public expression databases, CK+, Oulu-CASIA, TFD, and SFEW demonstrate that our method achieves better results than state-of-the-art. 
\end{abstract}

%%%%%%%%%%%%%%%%%%%%%%%%%%%%%%%%%%%%%%%%%%%%%%%%%%%%%%%%%%%%%%%%%%%%%%%%%%%%%%%%
\section{INTRODUCTION}
%big dataset is important to deep learning
Deep Convolutional Neural Networks (DCNN) have demonstrated impressive performance improvements for many problems in computer vision. One of the most important reasons behind its success is the availability of large-scale training databases, for example, ImageNet~\cite{imagenet_cvpr09} for image classification, Places~\cite{zhou2014learning} for scene recognition, CompCars~\cite{yang2015large} for fine-grained recognition and MegaFace~\cite{kemelmacher2015megaface} for face recognition. 

However, it is not uncommon to have small datasets in many application areas, facial expression recognition being one of them. With a relatively small set of training images, even when regularization techniques such as Dropout~\cite{srivastava2014dropout} and Batch Normalization~\cite{ioffe2015batch} are used, the results are not satisfactory. The mostly used method is to fine-tune a network that has been pre-trained on a large dataset. Because of the generality of the pre-learned features, this approach has achieved great success~\cite{girshick2014rich}. 

Motivated by this observation, several previous works~\cite{levi2015emotion, zhao2016peak} on expression recognition utilize face recognition datasets to pre-train the network, which is then fine-tuned on the expression dataset.  
%It has been observed that the two tasks are closely related~\cite{zhong2012learning}. Moreover, 
The large amount of labeled face data~\cite{kemelmacher2015megaface, yi2014learning}, makes it possible to train a fairly complicated and deep network. Moreover, the close relationship between the two domains facilites the transfer learning of features.

\begin{figure}[!ht]
  \centering
    \includegraphics[scale=0.3]{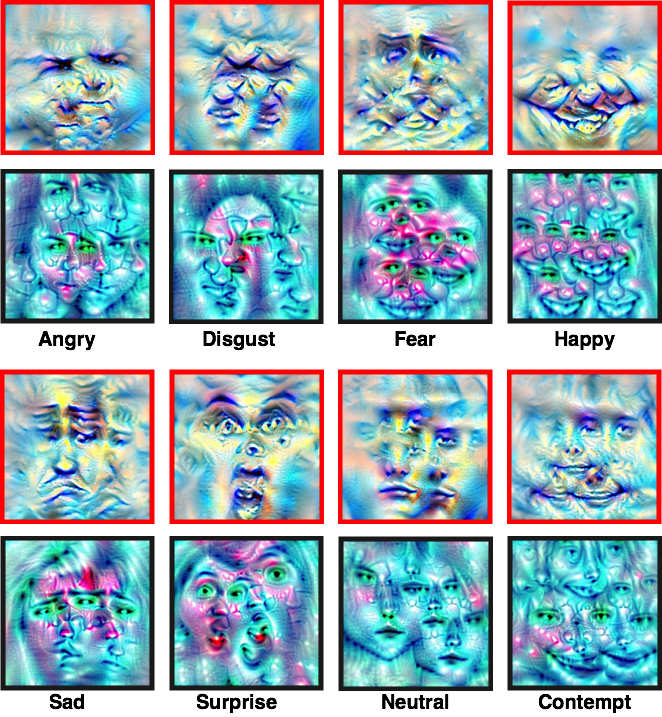}
  \caption{The red-boxed images are generated by the model trained with our method, while the black-boxed images are from the face network fine-tuned on the expression dataset. We can see the images produced by the face net are dominated with faces, while our model represents the facial expressions better. Models are visualized by DeepDraw~\cite{deepdraw}. }
  %More visualizations can be found in supplementary file.}
  \label{figurelabel}
  \vspace{-2mm}
\end{figure}

Although this strategy performs well, it has two notable problems: (i) the fine-tuned face net may still contain information useful for subject identification. This is because of the large size gap (several orders of magnititudes) between face and expression datasets.
%and this information may still exist even after fine-tuning. because of the big capacity of the network and the large gap between the size of face and expression datasets, there are maybe still some information related to subject ids left even after fine-tuning.
%because the face net is well trained on the face dataset, even after fine-tuning, there are still much 
%there are still much information useful for face identification left. 
As we can see from Fig. 1, the images (black-boxed) generated by the face net are dominated by faces as they should, which weakens the network's ability to represent the different expressions. 
%there are many faces in the generated images (black boxed), and the learned expression semantics are not very clear; second, because the network is pre-trained on a large dataset, to have good performance, it is usually very deep and big. 
(ii) the network designed for the face recognition domain is often too big for the expression task, thus the overfitting issue is still severe. 
%On the other hand, the network architecture is overcomplicated, which severs the overfitting issue. As the face dataset is usually very large, the designed network is very deep. For the small emotion dataset, ideally we
%this method doesn't give us the freedom to design the network. 
%the expression dataset is fairly small. This brings two issues: sever overfitting and over-complicated network architecture. In a recent work~\cite{sikka2016lomo}, the author found that the features from face network performs even worse than traditional features like SIFT. Thus, how to utilize the face recognition task for expression recognition is still an open question.

In this paper, we present \textbf{FaceNet2ExpNet}, a novel learning algorithm that incorporates face domain knowledge to regularize the training of an expression recognition network. 
Specially we first propose a new distribution function to model the high-level neurons of the expression net using the information derived from the fine-tuned face net. 
Such modeling naturally leads to a regression loss which serves as feature-level regularization that pushes the intermediate features of the expression net to be close to those of the fine-tuned face net. Next, to further improve the discriminativeness of the learned features, we refine the network with strong supervision from the label information. 
%While at the same time,  we still allow it to refine the learned features to be more discriminative. We design our network following the traditional convolutional plus fully-connected architecture.
We adopt a conventional network architecture, consisting of convolutional blocks followed by fully-connected layers, to design our expression net.
The training is carried out in two stages: 
in the first stage, only the convolutional layers are trained. We utilize the deep features from the face net as the supervision signal to make the learning easier. It also contains meaningful knowledge about human faces, which is important for expression recognition, too. 
After the first stage of learning is completed,  we add randomly initialized fully-connected (FC) layers and jointly train the whole network using the label information in the second stage. As observed by previous works~\cite{vittayakorn2016automatic}, FC layers generally capture domain-specific semantics. So we only utilize the face net to guide the learning of the convolutional layers and the FC layers are trained from scratch. Moreover, we empirically find that late middle layer (\emph{e.g.} pool5 for VGG-16~\cite{Simonyan14c}) is more suitable for training supervision due to the richness of low entropy neurons.
%We also carry out extensive experiments to empirically validate this design choice in Section 3.
%This training scheme has the following benefits:
In both training stages, only expression images are used.
%we first propose a new exponentially decay distribution function for the output response maps. The mean is modeled by the face net. Based on this, we formulate the loss function of the first stage as a regression loss. Only the convolutional layers of the emotion net is trained in this stage, while the fine-tuned face net is frozen. The inputs are the expression images, and the label information is not used.
%It consists of two steps: in the first stage, we train the convolutional blocks of the expression network. The supervision comes from the last pooling layer of the face net, and we don't need the image label information. The used face net has already fine-tuned on the expression dataset,  and frozen in this stage.
%In the second stage, we attach fully connected layers which is randomly initialized to the trained convolutional blocks, and joint train the two parts together with cross-entropy loss. 

From Fig. 1, we can see that the models trained with our method capture the key properties of different expressions. For example, the angry expression is displayed by frowned eye brows and a closed mouth; the surprise expression is represented by a large opened mouth and eyes. 
%The face net acts as a regularizer that guides the emotion net learning by transferring the supervision. Since it is trained from scratch with only expression images, the emotion net captures more meaningful high-level semantics. As we can see in Fig. 1, the expression images produced by our model are very vivid. For example, the angry emotion is displayed by frowned eye brows and closed mouth; the surprise emotion is represented by large opened mouth and eyes. 
This method is different from knowledge distillation~\cite{hinton2015distilling}. Here we do not have a large accurate network trained on the same domain to produce reliable outputs from softmax. It is also different from FitNets~\cite{romero2014fitnets}, which is mainly used to train a thinner and deeper network.

%To our knowledge, this is the first work that leverages a face net to  help train an expression net. 
%Our main \textbf{contributions} are 1) A new training  algorithm that is specially designed for expression recognition; 2)  
To validate the effectiveness of our method, we perform experiments on both constrained (CK+, Oulu-CASIA, TFD) and unconstrained datasets (SFEW). For all the four datasets, we achieve better results than the current state-of-the-art.  

The remainder of this paper is organized as follows. Section 2 briefly introduces related works. The FaceNet2ExpNet algorithm is presented in Section 3. Experiments and computational analysis are discussed in Section 4 and Section 5. We conclude this work in Section 6.

%however, in some area, it is hard to get a large dataset

%one of most used methods is fine-tuning. like r-cnn. since fr and ef are closely related, it's natual.

%limitation of fine-tune

%in this paper, introduce methods

%wrap up: stress contribution, stress experiment results

\section{Related Works}
In~\cite{zhong2012learning}, Zhong et al. observed that only a few active facial patches are useful for expression recognition. These active patches include: common patches for the recognition of all expressions and specific patches that are only important for single expression. To locate these patches, a two-stage multi-task sparse learning framework is proposed. In the first stage, multi-task learning with group sparsity is performed to search for the common patches. In the second stage, face recognition is utilized to find the specific patches. However, the sequential search process is likely to find overlapped patches. To solve this problem, Liu et al.~\cite{liu2014feature} integrated the sparse vector machine and multi-task learning into a unified framework. Instead of performing the patch selection in two separate phrases, an expression specific feature selection vector and a common feature selection vector are employed together. 
To get more discriminative features instead of hand-crafted features, Liu et al.~\cite{liu2013aware} used patch-based learning method. Subsequently, a group feature selection scheme based on the maximal mutual information and minimal redundancy criterion is presented. 
Lastly, three layers of restricted Boltzman machines (RBM) are stacked to learn hierarchical features.
%This is motivated by the domain knowledge that facial expression can be decomposed into a combination of facial action units (AU).  Interestingly, \cite{khorrami2015deep} also showed that different neurons in the convolutional layers resemble facial AUs. 
To further boost the performance, a loopy boosted deep belief network (DBN) framework was explored in~\cite{liu2014facial}. Feature learning, feature selection and classifier design are learned jointly. In the forward phase, several DBNs extract features from the overlapped facial patches. Then, AdaBoosting is adopted to combine these patch-based DBNs. In the fine-tuning phase, the loss from both weak and strong classifiers are backproped. 
In~\cite{liu2014learning}, to utilize the temporal information for video-based expression recognition, 3D CNN was applied to learn low-level features. Then, a GMM model is trained on the features, and the covariance matrix for each component composes the expressionlet. 
Motivated by the domain knowledge that facial expression can be decomposed into a combination of facial action units (AU), a deformable facial part model was explored in~\cite{liu2014deeply}. 
Multiple part filters are learned to detect the location of discriminative facial parts. To further cope with the pose and identity variations, a quadratic deformation cost is used.

More recently, Jung et al.~\cite{jung2015deep} trained a deep temporal geometry network and a deep temporal appearance network with facial landmarks and images. To effectively fuse these two networks, a joint fine-tuning method is proposed. Specifically, the weight values are frozen and only the top layers are trained. 
In~\cite{mollahosseini2016going}, Mollahosseini et al. discovered that the inception network architecture works very well for expression recognition task. Multiple cross dataset experiments are performed to show the generality of the learned model. In~\cite{yu2015image, ng2015deep}, a two-step training procedure is suggested, where in the first step, the network was trained using a relatively large expression dataset followed by training on the target dataset. Even though the image is of low resolution and the label of the relatively large dataset is noisy, this approach is effective. The work closely related to ours is~\cite{zhao2016peak}, which proposed to employ a peak expression image (easy sample) to help the training  of a network with input from a weak expression image (hard sample). This is also achieved by a regression loss between the intermediate feature maps. However, a pair of the same subject and the same expression image is required as input for training. This is not always possible, especially in unconstrained expression recognition scenario, where the subject identities are usually unknown.  

\section{Approach}
\subsection{Motivation}
We write our expression net as:
$$O = h_{\theta_2}(g_{\theta_1}(I))$$
where $h$ represents the fully connected layers, and $g$ corresponds to the convolutional layers. $\theta_2$ and $\theta_1$ are the parameters to be learned. I is the input image, and O is the output before softmax.

First, the parameters $\theta_1$ of the convolutional layers are learned.
%It is discovered in many previous works~\cite{} that  features from the fully connected layers are very discriminative, thus it is better to train them from scratch on the emotion datasets. 
%While for the convolutional blocks, we leverage the knowledge from the face net to regularize the emotion net.
In~\cite{xie2016interactive}, Xie et al. observed that the high-level neurons are exponentially decayed. To be more specific, by denoting the outputs of the $l_{th}$ layer as $x_{c,w,h}$, 
and the average response value over the spatial dimension as
\begin{equation}
x_c = \frac{1}{W \times H}\sum_{w=0}^{W-1}\sum_{h=0}^{H-1}x_{c,w,h}
\end{equation}
where $C$ is the number of output channels in the $l_{th}$ layer, and $W$, $H$ is the  width and height of the response maps, respectively.
Then the distribution function can be formulated as follows: 
\begin{equation}
f(X^l) = C_p \cdot e^{-||X^l||_p^p}
\end{equation}
where $X^l = [x_1, ..., x_C] \in R^C$, and $C_p$ is a normalization constant. $||\cdot||_p^p$ is the $p_{th}$ norm. 

To incorporate the knowledge of a face net, we propose to extend (2) to have the following form, \emph{i.e.}, :
\begin{equation}
f(X^l) = C_p \cdot e^{-||X^l - \mu||_p^p}
\end{equation}
The mean is modeled by the face net, $\mu = G(I)$.  And $G$ represents the face net's convolutional layers. 
This is motivated by the observation that the fine-tuned face net already achieves competitive performance on the expression dataset, so it should provide a good initialization point for the expression net. Thus, we do not want the latter to deviate much from the former. 
% Different from (2), the outputs of our emotion net have non-zero mean, and is provided by the face net. Since the face net is well-trained due to the large-scale face identification dataset, it gives a good initialization for the emotion net. Moreover, face recognition and expression recognition are two closely-related tasks, the knowledge learnt by face net should also be useful for the emotion net. So we want the output of the emotion net not too far away from the face net. 
%Note, if using the standard training method, then (3) reduces to (2).

Using the maximum likelihood estimation (MLE) procedure, we can derive the loss function as:
\begin{align}
\begin{split}
&\max_{\theta_{1}} L_1 = \max_{\theta_{1}} \log f(X^l) \\
&=\max_{\theta_{1}} \log C_p \cdot e^{-||X^l - \mu}||\\
&=\min_{\theta_{1}} ||g_{\theta_1}(I) - G(I)||_p^p\\
\end{split}
\end{align}
Note that if $p=2$ and without $G$, this is the normal $l_2$ regularizer. Thus we can also view the face net acting as a regularizer, which stabilizes the training step of the expression net.
%In the regression loss, the parameters of the face net is fixed. The outputs from the intermediate layers are served as supervision to train the emotion net.  

\begin{figure}[!ht]
  \centering
  \includegraphics[width=0.5\textwidth]{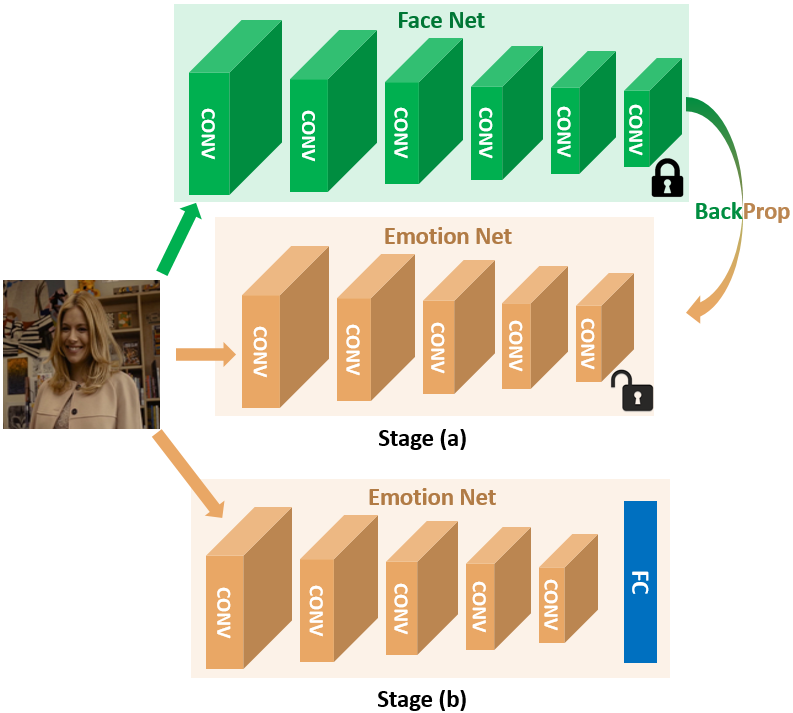}
  \caption{Two-stage Training Algorithm. In stage (a), the face net is frozen and provides supervision for the expression net. The regression loss is backproped only to the expression net. The convolutional layers are trained in this stage. In stage (b), the randomly initialized fully-connected layers are attached to the trained convolutional blocks. The whole network is trained jointly with cross-entropy loss. The face net is normally much deeper than the expression net.}
  \label{figurelabel}
\end{figure}

\subsection{Training Algorithm}
The training algorithm has the following two steps:

In the first stage, we train the convolutional layers with the loss function in (4). The face net is frozen, and the outputs from the last pooling layer are used to provide supervision for the expression net. We provide more explanations on this choice in the next section.
%We choose this layer because it is observed by several works~\cite{vittayakorn2016automatic} that deeper layer captures high-level semantics, which is important for expression recognition. 

In the second stage, we append the fully connected layers to the trained convolutional layers. The whole network is jointly learned using the cross-entropy loss, defined as follows:
\begin{equation}
L_2 = - \sum_{i=1}^N \sum_{j=1}^M y_{i,j} \log \hat{y}_{i,j}
\end{equation}
Where $y_{i,j}$ is the ground truth for the image, and $\hat{y}_{i,j}$ is the predicated label.
 The complete training algorithm is illustrated in Fig. 2.

Our expression net consists of five convolutional layers, each followed by a non-linear activation function (ReLU)  and a max-pooling layer. The kernel size of all the convolutional layers is a $3 \times 3$ window. For the pooling layer, it is $3 \times 3$ with stride 2. The numbers of the output channels are 64, 128, 256, 512, 512. After the last pooling layer, we add another $1 \times 1$ convolutional layer, which serves to bridge the gap between face and expression domains. Moreover, it also helps to adapt the dimension if the last pooling layer of the expression net does not match the face net.
To reduce overfitting, we have only one fully-connected layer with dimension 256. 
Note, if the spatial size of the last pooling layer between the face net and expression net does not match exactly, then deconvolution (fractionally strided convolution) can be used for upsampling.

%\begin{figure}[!ht]
%  \centering
%  \includegraphics[scale=0.3]{oulu_mean_image.png}
%  \caption{Confusion Matrix of CK+ for the Eight Class problem. The darker the color, the higher the accuracy.}
%  \label{figurelabel}
%  \vspace{-2mm}
%\end{figure}

\subsection{Which Layer to Transfer?}
In this section, we explore the layer selection problem for the first stage supervision transfer. Since the fine-tuned face network outperforms the pre-trained network on expression recognition, we hypothesize that there may be interesting differences in the network before and after fine-tuning. These differences might help us understand better which layer is more suitable to guide the training of the expression network. 

%low entropy is good
To this end, we first investigate the expression sensitivity of the neurons in the network, using VGG-16 as a working example. For each neuron, the images are ranked by the maximum response values. Then the top $K$ ($K$ = 100 in our experiments) images are binned according to the expression labels. We compute the entropy for the neuron  $x$ as $H(x) = - \sum_{i=1}^n p(i) \log p(i)$, where $p(i)$ denotes the histogram count for bin $i$ and $n$ denotes the number of quantized label bins (we normalize the histogram to sum to 1). If the neuron has a low entropy, then it should be more expression sensitive since its label distribution is peaky. To validate our assumption, we histogram the entropy for pool4, pool5, FC6 and FC7 layers. In Fig. 3, it is interesting to notice that as the layer goes deeper, more low entropy neurons start to emerge in the fine-tuned network compared with the pre-trained network. This phenomenon is particularly obvious in the fully-connected layers, which are often treated as discriminative features. While for pool4, the distribution does not change too much. 

%low entropy is starting to increase from pool5
Since these low entropy neurons indicate layer discriminativeness, we next compute the number of low expressive score (LES) neurons for each layer (here low expressive score is the entropy lower than the minimum average entropy score among the four selected layers). In Table I., we find that in comparison with the pre-trained network, the LES neurons increase dramatically in the fine-tuned network, especially starting from pool5 layer. Moreover, convolutional layers have a larger number of these neurons than FC layers. 
These results suggest that maybe late middle layer, such as pool5, is a good tradeoff between supervision richness and representation discriminativeness.

\begin{figure*}
  \includegraphics[width=\textwidth,height=7cm]{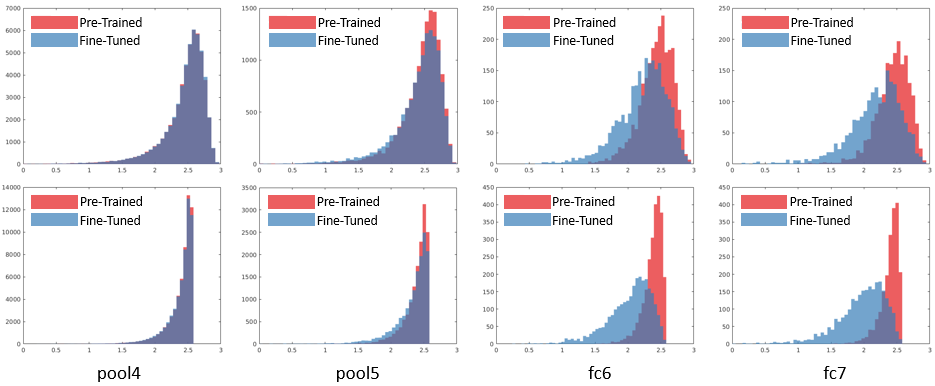}
  \caption{Histograms of neuron entropy scores from four different layers for pre-trained network (red) and fine-tuned network (blue). The X axis is the entropy value and the Y axis is the number of neurons. The first row is on CK+ dataset, while the second row is on Oulu-CASIA dataset.}
  \label{vgg_vs_ftvgg}
  \vspace{-1mm}
\end{figure*}

%\begin{table}
%\caption{the expressive score of CK+ \& Oulu-CASIA, the lower the better}
%\label{table_example}
%\begin{center}
%\begin{tabular}{|c||c||c||c||c|}
%\hline
% Model &  Pool4 & Pool5 & FC6 & FC7  \\
%\hline
%\hline
%Pre-trained & 2.4420 &2.4412  & 2.4302  & 2.4380  \\
%Fine-tuned & +0.0029  & -0.0475  & -0.2123  & -0.2715  \\
%\hline
%Pre-trained& 7763 & 2011 & 338 & 248\\
%Fine-tuned & -57 & +511 & +658 & +610\\
%\hline
%\hline
%Pre-trained  & 2.3849 & 2.3830 & 2.3788 & 2.3911 \\
%Fine-tuned & -0.0072 & -0.0755 & -0.3541 & -0.4241 \\
%\hline
%Pre-trained  & 3009 & 605 & 48 & 33 \\
%Fine-tuned & +194 & +895 & +952 & +1086 \\
%\hline
%\end{tabular}
%\end{center}
%\vspace{-1mm}
%\end{table}

\begin{table}
\caption{the number of low expressive score neurons for pre-trained network and fine-tuned network}
\label{table_example}
\begin{center}
\begin{tabular}{|c||c||c||c||c|}
\hline
 Model &  Pool4 & Pool5 & FC6 & FC7  \\
\hline
Pre-trained (CK)& 7763 & 2011 & 338 & 248\\
Fine-tuned (CK)& -57 & +511 & +658 & +610\\
\hline
Pre-trained  (Oulu-CASIA)& 3009 & 605 & 48 & 33 \\
Fine-tuned (Oulu-CASIA)& +194 & +895 & +952 & +1086 \\
\hline
\end{tabular}
\end{center}
\vspace{-1mm}
\end{table}

\section{Experiments}
We validate the effectiveness of our method on four widely used databases: CK+~\cite{lucey2010extended}, Oulu-CASIA~\cite{zhao2011facial}, Toronto Face Database (TFD)~\cite{susskind2010toronto} and Static Facial Expression in the Wild (SFEW)~\cite{dhall2015video}.  The numbers of images for different expressions are shown in Table. II. 
%The faces in the first two datasets are mostly frontal, while the images in the last one are extracted from films with large poses. 
In the following, we reference our method FaceNet2ExpNet as FN2EN.
\begin{table}
\caption{the number of images for different expression classes}
\label{table_example}
\begin{center}
\resizebox{\columnwidth}{!}{
\begin{tabular}{|c||cccccccc||c|}
\hline
&An & Co & Di & Fe & Ha & Sa & Su & Ne & Total\\
\hline
CK+ & 135 & 54 & 177 & 75 & 147 & 84 & 249 & 327 & 1308\\
Oulu-CASIA & 240 & & 240 & 240 & 240 & 240 & 240 &  &1444\\
TFD         & 437 &    & 457 & 424 & 758  & 441 & 459  & 1202   &4178\\
SFEW & 255 & & 75 & 124 & 256 & 234 & 150 & 228 & 1322 \\
\hline
\end{tabular}
}
\vspace{-1mm}
\end{center}
\end{table}

\begin{figure}[!ht]
  \centering
  \includegraphics[width=0.5\textwidth]{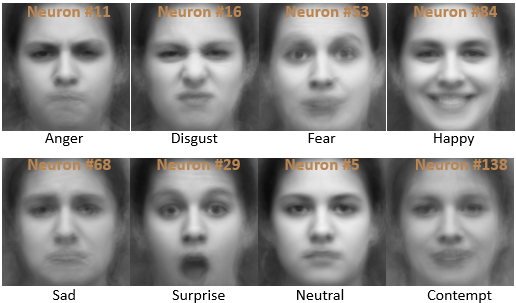}
  \caption{Visualizes several neurons in the top hidden layer of our model on CK+ dataset.}
  \label{figurelabel}
  \vspace{-2mm}
\end{figure}

\begin{figure}[!ht]
  \centering
  \includegraphics[scale=0.25]{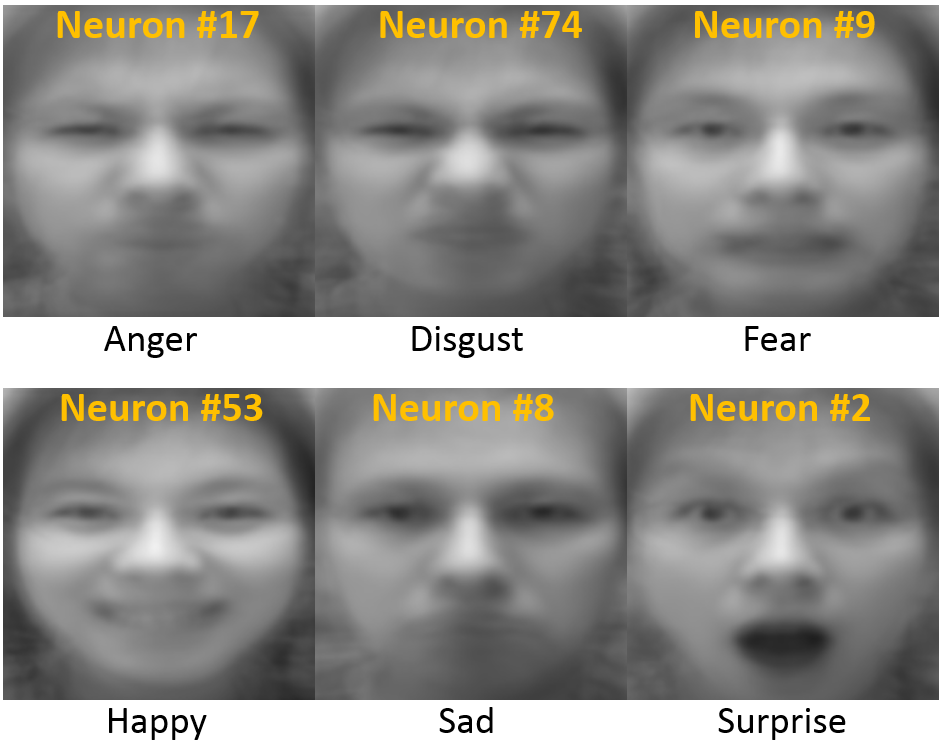}
  \caption{Visualizes several neurons in the top hidden layer of our model on Oulu-CASIA dataset.}
  \label{figurelabel}
  \vspace{-2mm}
\end{figure}

\subsection{Implementation}
We apply the Viola Jones~\cite{viola2004robust} face detector and IntraFace~\cite{de2015intraface} for face detection and landmark detection. The faces are normalized, cropped, and resized to $256 \times 256$. We utilize conventional data augmentation in the form of random sampling and horizontal flipping. The min-batch size is 64, the momentum is fixed to be 0.9 and the dropout is set at 0.5.

For network training, in the first stage, the regression loss is very large. So we start with a very small learning rate 1e-7, and decrease it after 100 epochs. The total training epochs for this stage is 300. We also try gradient clipping, and find that though it enables us to use a bigger learning rate, the results are not better compared to when a small learning rate was used. In the second stage, the fully connected layer is randomly initialized  from a Gaussian distribution, and the convolutional layers are initialized from the first stage. The learning rate is 1e-4, and decreased by 0.1 after 20 epochs. We train it for 50 epochs in total. Stochastic Gradient Descent (SGD) is adopted as the optimization algorithm. For testing, a \textbf{single center crop} with size  $224 \times 224$ is used. The settings are same for all the experiments. We use the face net from~\cite{parkhi2015deep}, which is trained on 2.6M face images. All the experiments are performed using the deep learning framework Caffe~\cite{jia2014caffe}. Upon publication, the trained expression models will be made publicly available.

\subsection{Neuron Visualization}
We first show that the model trained with our algorithm captures the semantic concepts related to facial expression very well. Given a hidden neuron, the face images that obtain high response are averaged. We visualize these mean images for several neurons in Fig. 4 and Fig. 5 on CK+ and Oulu-CASIA, respectively. Human can easily assign each neuron with a semantic concept it measures (\textit{i.e.} the text in black). For example, the neuron 11 in the first column in Fig. 4 corresponds to \say{Anger}, and the neuron 53 in Fig. 5 represents \say{Happy}. Interestingly, the high-level concepts learned by the neurons across the two datasets are very consistent.

\begin{figure}[!ht]
  \centering
  \includegraphics[width=0.5\textwidth]{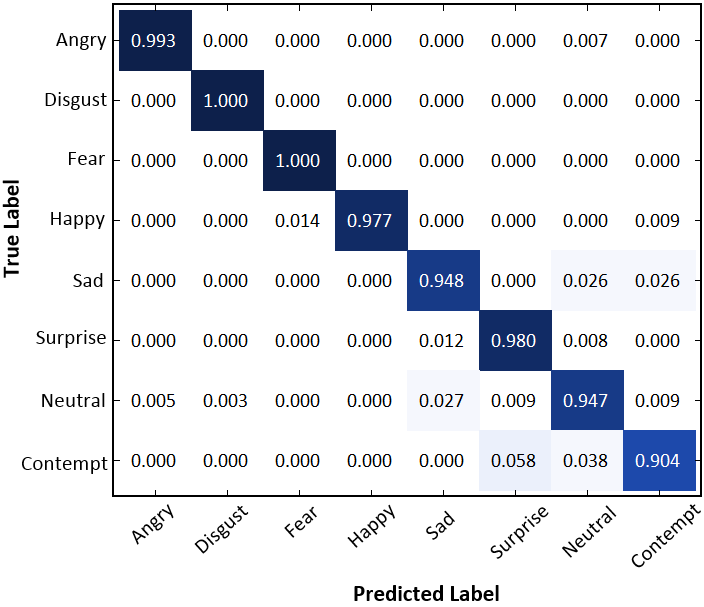}
  \caption{Confusion Matrix of CK+ for the Eight Classes problem. The darker the color, the higher the accuracy.}
  \label{figurelabel}
  \vspace{-2mm}
\end{figure}

\subsection{CK+}
CK+ consists of 529 videos from 123 subjects, 327 of them annotated with eight expression labels. Each video starts with a neutral expression, and reaches the peak in the last frame.  As in other works~\cite{liu2014learning}, we extract the last three frames and the first frame of each video to compose our image-based CK+ database. The total number of the images is 1308, which is split into 10 folds. The subjects are divided into ten groups by ID in ascending order. 

\begin{table}
\caption{The Average Accuracy on CK+ dataset}
\label{accuracy_ck}
\begin{center}
\begin{tabular}{|c||c||c|}
\hline
Method & Average Accuracy & \#Exp. Classes\\
\hline
CSPL~\cite{zhong2012learning} & 89.9\% & Six Classes\\
AdaGabor~\cite{bartlett2005recognizing} & 93.3\% &\\
LBPSVM~\cite{feng2007facial} & 95.1\% &\\
3DCNN-DAP~\cite{liu2014deeply} & 92.4\% &\\
BDBN~\cite{liu2014facial} & 96.7\% &\\
STM-ExpLet~\cite{liu2014learning} & 94.2\% &\\
DTAGN~\cite{jung2015deep} & 97.3\% &\\
Inception~\cite{mollahosseini2016going} & 93.2\% &\\
LOMo~\cite{sikka2016lomo} & 95.1\% &\\
PPDN~\cite{zhao2016peak} & 97.3\% &\\
FN2EN & \textbf{98.6\%} &\\
\hline
AUDN~\cite{liu2013aware} & 92.1\% & Eight Classes\\
Train From Scratch (BN) & 88.7\% &\\
VGG Fine-Tune (baseline) & 89.9\% &\\
FN2EN & \textbf{96.8\%} &\\
\hline
\end{tabular}
\end{center}
\vspace{-2mm}
\end{table}

In Table III, we compare our approach with both traditional and deep learning-based methods in terms of average accuracy. We consider the fine-tuned VGG-16 face net as our baseline. To further show the superiority of our method, we also include the results on training from scratch with batch normalization. The network architecture is same as FNEN. The first block shows the results for six classes, while the second block shows the results for eight classes, including both contempt and neutral expressions. Among them, 3DCNN-DAP~\cite{liu2014deeply}, STM-ExpLet~\cite{liu2014learning} and DTAGN~\cite{jung2015deep} are image-sequence based methods, while others are image-based. For both cases, our method significantly outperforms all others, achieving 98.6\% vs the pervious best of 97.3\% for six classes, and 96.8\% vs 92.1\% for eight classes.

Because of the high accuracy on the six class problem, here we only show the confusion matrix for eight class problem. From Fig. 6 we can see that both disgust and fear expressions are perfectly classified, while contempt is the most difficult to classify. It is because this expression has the least number of training images, and the way people show it is very subtle. 
Surprisingly, from the visualization in Fig. 1, the network is still able to capture the speciality of contempt: the conner of the mouth is pulled up. This demonstrates the effectiveness of our training method.

\begin{table}
\caption{The Average Accuracy on Oulu-CAS dataset}
\label{table_example}
\begin{center}
\begin{tabular}{|c||c|}
\hline
Method & Average Accuracy\\
\hline
HOG 3D~\cite{klaser2008spatio} & 70.63\%\\
AdaLBP~\cite{zhao2011facial} & 73.54\%\\
Atlases~\cite{guo2012dynamic} & 75.52\%\\
STM-ExpLet~\cite{liu2014learning} & 74.59\%\\
DTAGN~\cite{jung2015deep}  & 81.46\%\\
LOMo~\cite{sikka2016lomo} & 82.10\%\\
PPDN~\cite{zhao2016peak}& 84.59\%\\
Train From Scratch (BN) & 76.87\%\\
VGG Fine-Tune (baseline) & 83.26\%\\
\hline
FN2EN & \textbf{87.71\%}\\
\hline
\end{tabular}
\end{center}
\vspace{-2mm}
\end{table}

\subsection{Oulu-CAS VIS}
Oulu-CASIA has 480 image sequences taken under Dark, Strong, Weak illumination conditions. In this experiment, only videos with Strong condition captured by a VIS camera are used. There are 80 subjects and six expressions in total. Similar to CK+, the first frame is always neutral while the last frame has the peak expression. Only the last three frames are used, and the total number of images is 1440. A ten-fold cross validation is performed, and the split is subject independent.

\begin{figure}[!ht]
  \centering
  \includegraphics[width=0.5\textwidth]{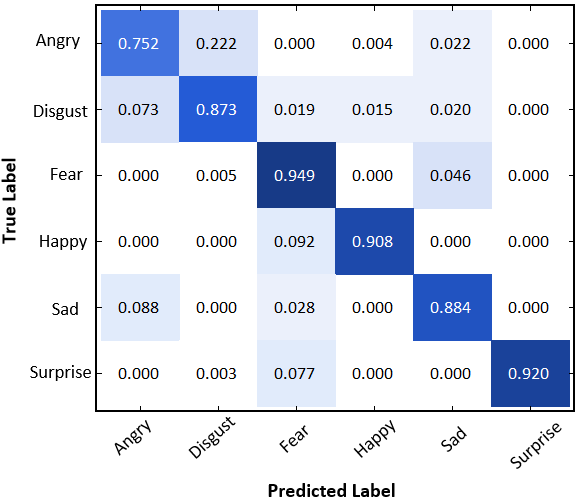}
  \caption{Confusion Matrix of Oulu-CASIA. The darker the color, the higher the accuracy.}
  \label{figurelabel}
  \vspace{-2mm}
\end{figure}

Table IV reports the results of average accuracy for the different approaches. As can be seen, our method achieves substantial improvements over the previous best performance achieved by PPDN~\cite{zhao2016peak}, with a gain of \textbf{3.1\%}. %Notice, the baseline is already very strong on this dataset. 
The confusion matrix is shown in Fig. 7. The proposed method performs well in recognizing fear and happy, while angry is the hardest expression, which is mostly confused with disgust. 
%We also visualize the learned model in Fig. 5. Consistent with the result from the confusion matrix, the images generated for angry and disgust classes are very similar.

\begin{table}
\caption{The Average Accuracy on TFD dataset}
\label{table_example}
\begin{center}
\begin{tabular}{|c||c|}
\hline
Method & Average Accuracy\\
\hline
Gabor + PCA~\cite{dailey2002empath} & 80.2\%\\
Deep mPoT~\cite{susskind2011deep} & 82.4\%\\
CDA+CCA~\cite{rifai2012disentangling} & 85.0\%\\
disRBM~\cite{reed2014learning} & 85.4\%\\
bootstrap-recon~\cite{reed2014training} & 86.8\%\\
Train From Scratch (BN) & 82.5\%\\
VGG Fine-Tune (baseline) & 86.7\%\\
\hline
FN2EN & \textbf{88.9\%}\\
\hline
\end{tabular}
\end{center}
\vspace{-2mm}
\end{table}
\subsection{TFD}
The TFD is the largest expression dataset so far, which is comprised of images from many different sources. It contains \emph{4178} images, each of which is assigned one of seven expression labels. The images are divided into 5 separate folds, each containing train, valid and test partitions. We train our networks using the training set and report the average results over five folds on the test sets.

Table V summarizes our TFD results. As we can see, the fine-tuned VGG face is a fairly strong baseline, which is almost on par with the current state-of-the-art, $86.7\%$ vs $86.8\%$. Our method performs the best, significantly outperforming bootstrap-recon~\cite{reed2014training} by $2\%$. From the confusion matrix, we find that fear has the lowest recognition rate and is easy to be confused with surprise. When inspecting the dataset, we find the images from the two expressions indeed have very similar facial appearances: mouth and eyes are wide open.

\begin{figure}[!ht]
  \centering
  \includegraphics[width=0.5\textwidth]{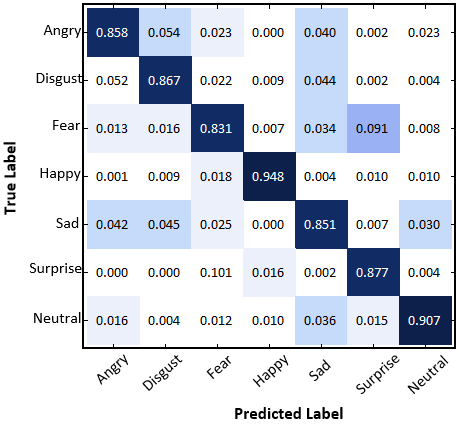}
  \caption{Confusion Matrix of TFD. The darker the color, the higher the accuracy.}
  \label{figurelabel}
  \vspace{-2mm}
\end{figure}

\subsection{SFEW}
Different from the previous three datasets, SFEW is targeted for unconstrained expression recognition. So the images are all extracted from films clips, and labeled with seven expressions. The poses are large, and the expression is much more difficult to recognize. Furthermore, it has only 891 training images. Because we do not have access to the test data, here we report the results on the validation data.

\begin{table}
\caption{The Average Accuracy on SFEW dataset}
\label{table_example}
\begin{center}
\begin{tabular}{|c||c||c|}
\hline
Method & Average Accuracy & Extra Train Data\\
\hline
AUDN~\cite{liu2013aware} & 26.14\% &None\\
STM-ExpLet~\cite{liu2014learning} & 31.73\% &\\
Inception~\cite{mollahosseini2016going} & 47.70\% & \\
Mapped LBP~\cite{levi2015emotion} & 41.92\% &\\
Train From Scratch (BN) &39.55\%&\\
VGG Fine-Tune (baseline) & 41.23\%&\\
FN2EN & \textbf{48.19\%} &\\
\hline
Transfer Learning~\cite{ng2015deep} & 48.50\% & FER2013\\
Multiple Deep Network~\cite{yu2015image} & 52.29\%&\\
FN2EN  & \textbf{55.15\%}&\\
\hline
\end{tabular}
\end{center}
\vspace{-2mm}
\end{table}

In Table VI, we divide the methods into two blocks, where the first block only uses the training images from SFEW, while the second block utilizes FER2013~\cite{goodfellow2013challenges} as additional training data.
For both settings, our method achieves best recognition rates. Especially with more training data, we surpass Multiple Deep Network Learning~\cite{yu2015image} by almost \textbf{3\%}, which is the runner-up in EmotiW 2015. We do not compare the result with the winner~\cite{kim2016hierarchical} since they use 216 deep CNNs to get 56.40\%, while we only use a single CNN (1.25\% higher than our method). From the confusion matrix Fig. 9, we can see the accuracy for fear is much lower than other expressions. This is also observed in other works~\cite{ng2015deep}. 
%In addition, we find in Fig. 6 that the images produced by the model trained on SFEW is less meaningful compared with CK+ and Oulu-CASIA. This is because the poses in SFEW are larger and the expressions are more nuanced.

\begin{figure}[!ht]
  \centering
  \includegraphics[width=0.5\textwidth]{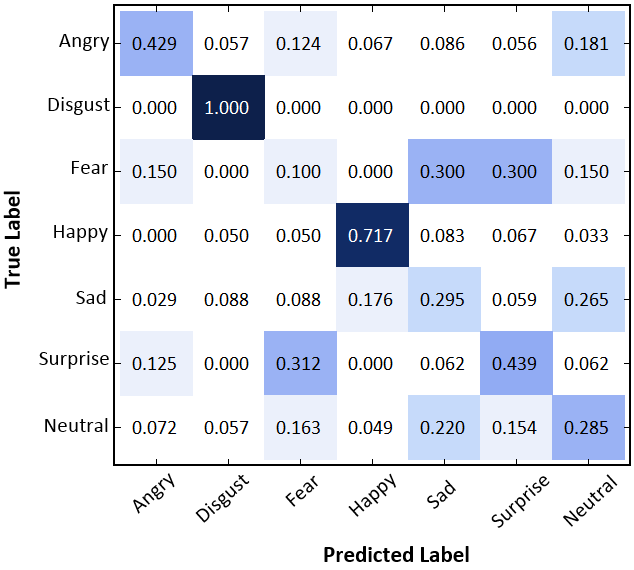}
  \caption{Confusion Matrix of SFEW. The darker the color, the higher the accuracy.}
  \label{figurelabel}
  \vspace{-1mm}
\end{figure}

%
%\begin{figure}[!ht]
%  \centering
%  \includegraphics[scale=0.3]{crop_oulu.png}
%  \caption{Model Visualization for Oulu-CASIA}
%  \label{figurelabel}
%  \vspace{-2mm}
%\end{figure}
%
%
%
%
%\begin{figure}[!ht]
%  \centering
%  \includegraphics[scale=0.3]{fer_sfew.png}
%  \caption{Model Visualization for SFEW}
%  \label{figurelabel}
%  \vspace{-7mm}
%\end{figure}

\section{Computational speed analysis}
Compared with networks adopted in previous works~\cite{ng2015deep, mollahosseini2016going, zhao2016peak}, AlexNet~\cite{krizhevsky2012imagenet} or  VGG-M~\cite{chatfield2014return}, the size of our network is fairly small. The number of parameters is 11M vs. VGG-16 baseline 138M. For testing, our approach takes only 3ms per image using a single Titan X GPU. 

%Compared with the networks adopted in previous works~\cite{ng2015deep, mollahosseini2016going, zhao2016peak}, AlexNet~\cite{krizhevsky2012imagenet}, GoogleNet~\cite{szegedy2015going} or  VGG-M~\cite{chatfield2014return}, the size of our network is fairly small (12M parameters). For testing, our approach takes only 3ms per image using a single Titan X GPU. 

% \begin{table}
%\caption{the Number of parameters and feature dimensions from different networks}
%\label{table_example}
%\begin{center}
%\begin{tabular}{|c||c||c|}
%\hline
%Method & \#Param.  &  Dim.\\
%\hline
%Mapped LBP~\cite{levi2015emotion} & 73M&2048\\
%Transfer Learning~\cite{ng2015deep} & 60M &4096\\
%Multiple Deep Network Learning~\cite{yu2015image} & 6M&1024\\
%VGG Fine-Tune (baseline) & 138M &4096\\
%\hline
%FN2EN & 11M&256\\
%\hline
%\end{tabular}
%\end{center}
%\end{table}
%PPDN~\cite{zhao2016peak} & 5M & 1024\\

%%%%%%%%%%%%%%%%%%%%%%%%%%%%%%%%%%%%%%%%%%%%%%%%%%%%%%%%%%%%%%%%%%%%%%%%%%%%%%%%
\section{CONCLUSIONS AND FUTURE WORKS}

%\subsection{Conclusions}

%In this paper, we present FaceNet2ExpNet, a novel two-stage training algorithm for expression recognition. 
%We first propose a probabilistic distribution function for the high-level neuron responses. Based on this, a two-stage training algorithm is proposed. We see that middle-level features of a face recognition network carries rich information to supervise the training of an expression net. FaceNet2ExpNet improves visual feature representation and outperforms various state-of-the-art methods on four public datasets. In future, we plan to apply this training method to other domains with small datasets.

In this paper, we present FaceNet2ExpNet, a novel two-stage training algorithm for expression recognition. 
In the first stage, we propose a probabilistic distribution function to model the high level neuron response based on already fine-tuned face net, thereby leading to feature level regularization that exploits the rich face information in the face net. In the second stage, we perform label supervision to boost the final discriminative capability. As a result, FaceNet2ExpNet improves visual feature representation and outperforms various state-of-the-art methods on four public datasets. In future, we plan to apply this training method to other domains with small datasets.

%In this paper, we propose to leverage a face net to regularize the training of an emotion net. This regularization not only helps transfer the related domain knowledge, but also alleviates the overfitting issue. Based on this, a two-stage training algorithm is designed. From the visualization, we can see the high-level semantics learnt by our method is more meaningful than pure fine-tune.  We achieve the current best results on three state-of-art facial expression benchmarks.
%\subsection{Future Works}
%A general extension of the weight-decayed function is the Mixture of Gaussian distribution. This formulation incorporates multiple related tasks into the training framework, instead of only one. Moreover, the training scheme is very general, and can also be applied to other domains with small datasets.

%%%%%%%%%%%%%%%%%%%%%%%%%%%%%%%%%%%%%%%%%%%%%%%%%%%%%%%%%%%%%%%%%%%%%%%%%%%%%%%%
\section{ACKNOWLEDGMENTS}
This research is based upon work supported by the Office of the Director of National Intelligence (ODNI), Intelligence Advanced Research Projects Activity (IARPA), via IARPA R\&D Contract No. 2014-14071600012. The views and conclusions contained herein are those of the authors and should not be interpreted as necessarily representing the official policies or endorsements, either expressed
or implied, of the ODNI, IARPA, or the U.S. Government. The U.S. Government is authorized to reproduce and distribute
reprints for Governmental purposes notwithstanding any copyright annotation thereon.
%
%The authors gratefully acknowledge the contribution of reviewers' comments, etc. (if desired). Put sponsor acknowledgments in the unnumbered footnote on the first page.

%%%%%%%%%%%%%%%%%%%%%%%%%%%%%%%%%%%%%%%%%%%%%%%%%%%%%%%%%%%%%%%%%%%%%%%%%%%%%%%%

%References are important to the reader; therefore, each citation must be complete and correct. If at all possible, references should be commonly available publications.

%\begin{thebibliography}{99}
%
%\bibitem{c1}
%J.G.F. Francis, The QR Transformation I, {\it Comput. J.}, vol. 4, 1961, pp 265-271.
%
%\bibitem{c2}
%H. Kwakernaak and R. Sivan, {\it Modern Signals and Systems}, Prentice Hall, Englewood Cliffs, NJ; 1991.
%
%\bibitem{c3}
%D. Boley and R. Maier, "A Parallel QR Algorithm for the Non-Symmetric Eigenvalue Algorithm", {\it in Third SIAM Conference on Applied Linear Algebra}, Madison, WI, 1988, pp. A20.
%
%\end{thebibliography}

{\small
\bibliographystyle{ieeetr}
\bibliography{egbib}
}

\end{document}